\documentclass[sigconf]{acmart}

\usepackage{graphicx}
\usepackage{epstopdf}
\usepackage{makecell}
\usepackage{booktabs}
\usepackage{multirow}
\usepackage{algorithm}
\usepackage{algorithmic}
\usepackage{subfigure}
\usepackage{enumitem}
\usepackage{enumerate}
\usepackage{amsthm}
\usepackage{soul}
\soulregister\caption7
\usepackage[framemethod=tikz]{mdframed}
\usepackage{lipsum}
\usepackage{tcolorbox}
\usepackage{balance}

\newtheorem{definition}{Definition}
\AtBeginDocument{%
	\providecommand\BibTeX{{%
			\normalfont B\kern-0.5em{\scshape i\kern-0.25em b}\kern-0.8em\TeX}}}

\copyrightyear{2021}
\acmYear{2021}
\setcopyright{acmcopyright}\acmConference[SIGMOD '21]{Proceedings of the 2021 International Conference on Management of Data}{June 18--27, 2021}{Virtual Event , China}
\acmBooktitle{Proceedings of the 2021 International Conference on Management of Data (SIGMOD '21), June 18--27, 2021, Virtual Event , China}
\acmPrice{15.00}
\acmDOI{10.1145/3448016.3452776}
\acmISBN{978-1-4503-8343-1/21/06}
\begin{document}
\fancyhead{}
\title{Joint Open Knowledge Base Canonicalization and Linking}
\newcommand\blfootnote[1]{%
	\begingroup
	\renewcommand\thefootnote{}\footnote{#1}%
	\addtocounter{footnote}{-1}%
	\endgroup
}

\author{Yinan Liu$^\dag$,$\ \;$Wei Shen$^\dag$$^*$,$\ \;$Yuanfei Wang$^\dag$,$\ \;$Jianyong Wang$^\ddag$,$\ \;$Zhenglu Yang$^\dag$, $\ \;$Xiaojie Yuan$^\dag$}
\affiliation{
\country{$^\dag$TKLNDST, College of Computer Science, Nankai University, Tianjin 300071, China}\\
\country{$^\ddag$Department of Computer Science and Technology, Tsinghua University, Beijing 100084, China} \\
	}
%\country{$^\ddag$Jiangsu Collaborative Innovation Center for Language Ability, Jiangsu Normal University, Xuzhou 221009, China
%}
\email{{liuyn, 1710270}@mail.nankai.edu.cn, {shenwei, yangzl, yuanxj}@nankai.edu.cn, jianyong@tsinghua.edu.cn}

\begin{abstract}
%	\vspace{-0.5mm}
	Open Information Extraction (OIE) methods extract a large number of OIE triples (noun phrase, relation phrase, noun phrase) from text, which compose large Open Knowledge Bases (OKBs). However, noun phrases (NPs) and relation phrases (RPs) in OKBs are not canonicalized and often appear in different paraphrased textual variants, which leads to redundant and ambiguous facts. To address this problem, there are two related tasks: OKB canonicalization (i.e., convert NPs and RPs to canonicalized form) and OKB linking (i.e., link NPs and RPs with their corresponding entities and relations in a curated Knowledge Base (e.g., DBPedia). These two tasks are tightly coupled, and one task can benefit significantly from the other. However, they have been studied in isolation so far. In this paper, we explore the task of joint OKB canonicalization and linking for the first time, and propose a novel framework JOCL based on factor graph model to make them reinforce each other. JOCL is flexible enough to combine different signals from both tasks, and able to extend to fit any new signals. A thorough experimental study over two large scale OIE triple data sets shows that our framework outperforms all the baseline methods for the task of OKB canonicalization (OKB linking) in terms of average F1 (accuracy).
\end{abstract}



%\begin{CCSXML}
%	<ccs2012>
%	<concept>
%	<concept_id>10002951.10002952.10003219.10003223</concept_id>
%	<concept_desc>Information systems~Entity resolution</concept_desc>
%	<concept_significance>500</concept_significance>
%	</concept>
%	<concept>
%	<concept_id>10002951.10003227.10003351.10003218</concept_id>
%	<concept_desc>Information systems~Data cleaning</concept_desc>
%	<concept_significance>500</concept_significance>
%	</concept>
%	</ccs2012>
%\end{CCSXML}
%
%\ccsdesc[500]{Information systems~Entity resolution}
%\ccsdesc[500]{Information systems~Data cleaning}

%\begin{CCSXML}
%	<ccs2012>
%	<concept>
%	<concept_id>10002951.10002952.10003219.10003218</concept_id>
%	<concept_desc>Information systems~Data cleaning</concept_desc>
%	<concept_significance>500</concept_significance>
%	</concept>
%	<concept>
%	<concept_id>10002951.10002952.10003219.10003223</concept_id>
%	<concept_desc>Information systems~Entity resolution</concept_desc>
%	<concept_significance>500</concept_significance>
%	</concept>
%	<concept>
%	<concept_id>10002951.10003227.10003351.10003218</concept_id>
%	<concept_desc>Information systems~Data cleaning</concept_desc>
%	<concept_significance>500</concept_significance>
%	</concept>
%	</ccs2012>
%\end{CCSXML}
%
%\ccsdesc[500]{Information systems~Data cleaning}
%\ccsdesc[500]{Information systems~Entity resolution}
%\ccsdesc[500]{Information systems~Data cleaning}

\maketitle
%\begin{CCSXML}
%	<ccs2012>
%	<concept>
%	<concept_id>10010520.10010553.10010562</concept_id>
%	<concept_desc>Computer systems organization~Embedded systems</concept_desc>
%	<concept_significance>500</concept_significance>
%	</concept>
%	<concept>
%	<concept_id>10010520.10010575.10010755</concept_id>
%	<concept_desc>Computer systems organization~Redundancy</concept_desc>
%	<concept_significance>300</concept_significance>
%	</concept>
%	<concept>
%	<concept_id>10010520.10010553.10010554</concept_id>
%	<concept_desc>Computer systems organization~Robotics</concept_desc>
%	<concept_significance>100</concept_significance>
%	</concept>
%	<concept>
%	<concept_id>10003033.10003083.10003095</concept_id>
%	<concept_desc>Networks~Network reliability</concept_desc>
%	<concept_significance>100</concept_significance>
%	</concept>
%	</ccs2012>
%\end{CCSXML}

%\ccsdesc[500]{Computer systems organization~Embedded systems}

%\vspace{-1mm}

\blfootnote{$^*$Corresponding author}
\blfootnote{$^\ddag$Also with Jiangsu Collaborative Innovation Center for Language Ability, Jiangsu Normal University, Xuzhou 221009, China.}


	\section{Introduction}
%	\vspace{-0.5mm}
	In recent years, several large curated Knowledge Bases (CKBs) with an ontology of pre-specified categories and relations have been developed, such as Freebase \cite{bollacker2008freebase}, DBpedia \cite{lehmann2015dbpedia}, and YAGO \cite{yago}. These CKBs contain millions of entities and hundreds of millions of relational facts between entities. In these CKBs, each entity is canonicalized and well defined with a unique identifier. CKBs play a key role for a variety of applications in both industry and academia. However, they are far from complete. As the world evolves, new entities and facts are generated. Enriching existing CKBs with external resources \cite{cao2020open, getman2018laying, min2017probabilistic, kruit2019extracting, yu2019knowmore} becomes more and more important.
	
\begin{figure*}[htbp] %需要subfigure宏包
		\subfigcapskip=-10pt
		\begin{minipage}[t]{\textwidth}
			\begin{minipage}[t]{\textwidth}
				\begin{figure}[H]
					\centering
					\subfigure[An illustration for the task of joint OKB canonicalization and linking.]{    %修改图一小标题
						\begin{minipage}[h]{.41\textwidth}
							\centering
							\includegraphics[width=\textwidth]{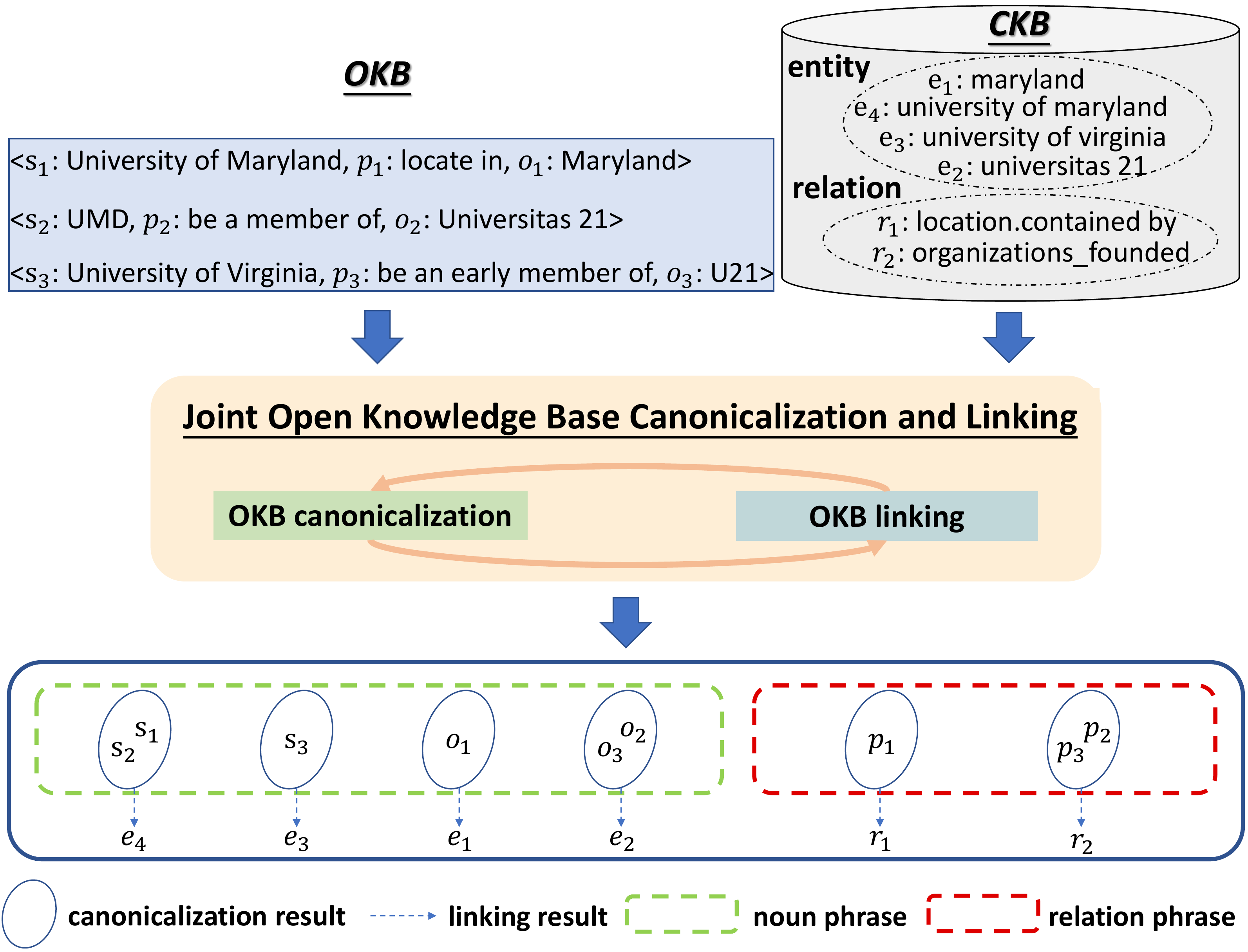}
							\label{task}
						\end{minipage}
					}
					\subfigure[A summarization of the proposed framework JOCL.]{    %修改图二小标题
						\begin{minipage}[h]{.56\textwidth}
							\centering
							\includegraphics[width=1.03\textwidth]{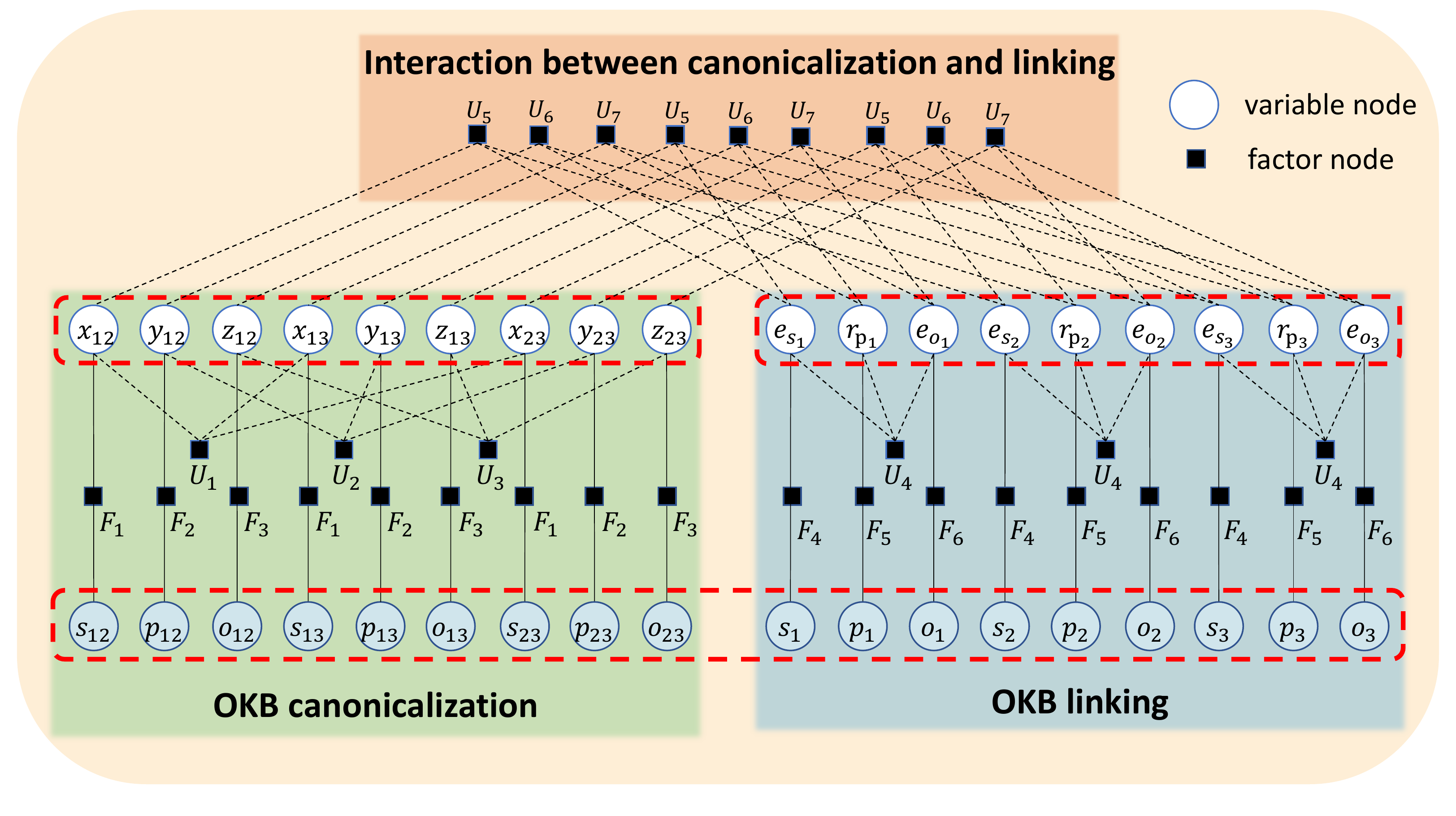}
							\label{framework}
						\end{minipage}
					}
					\caption{Task illustration and framework summarization.}
				\end{figure}
			\end{minipage}
		\end{minipage}
	\end{figure*}

	%\begin{figure}[t]
	%	\centering
	%	\includegraphics[width=3.22in]{task}
	%	\vspace{-1mm}
	%	\caption{An illustration for the task of joint OKB canonicalization and linking.}
	%	\label{task}
	%	\vspace{-6.2mm}
	%\end{figure}
	Open Information Extraction (OIE) methods without any pre-specified ontology can extract OIE triples of the form (noun phrase, relation phrase, noun phrase) from unstructured text documents. These large number of OIE triples compose large Open Knowledge Bases (OKBs), such as ReVerb \cite{fader2011identifying}, TextRunner \cite{Banko2007Open}, and OLLIE \cite{christensen2011analysis}. Unlike most CKBs confined to some encyclopedic knowledge sources (e.g., Wikipedia), the advantage of OKBs is that the coverage and diversity are much higher. Therefore, integrating OIE triples to CKBs is a significant and promising way for enriching existing CKBs \cite{dutta2015enriching}. However, compared with CKBs, OKBs are noisier and often plagued with ambiguity due to the lack of unique identifiers for the noun phrases and relation phrases in OIE triples. As the example shown in Figure \ref{task}, we list three OIE triples of an OKB. It can be seen that ``University of Maryland'' (i.e., $s_1$) and ``UMD'' (i.e., $s_2$) which are two noun phrases from different OIE triples refer to the same entity ``university of maryland'' in a CKB, and ``be a member of'' (i.e., $p_2$) and ``be an early member of'' (i.e., $p_3$) are two relation phrases from different OIE triples with the same semantic meaning which can be mapped to the same relation ``organizations\_founded'' in a CKB. To eliminate the ambiguity in OKBs and enrich CKBs with OIE triples, OKB canonicalization and OKB linking are two important tasks that need to be solved urgently.

	OKB canonicalization is the task of converting OIE triples of OKBs to canonicalized form, where noun phrases or relation phrases with the same semantic meaning are clustered to a group. Some models based on string similarity and embedding techniques \cite{galarraga2014canonicalizing, vashishth2018cesi, wu2018towards} have been proposed to canonicalize OKBs. A recent work \cite{lin2019canonicalization} achieved better performance by leveraging side information from the original source text.
	
	OKB linking is the task to jointly link noun phrases and relation phrases in OIE triples, with their corresponding real world entities and relations in a CKB. Traditional joint entity and relation linking methods for text \cite{dubey2018earl, sakor2019old,lin2020kbpearl} perform poorly on OIE triples with limited context, which has been confirmed by our experiments.

	From the two task definitions above, it can be seen that OKB canonicalization and OKB linking are closely related tasks. However, these two tasks have been studied in isolation so far. A heuristic way to integrate them is utilizing pipeline architecture, e.g., firstly canonicalizing OIE triples by an OKB canonicalization method, and then leveraging its output groups of noun phrases and relation phrases as the input for OKB linking. Unfortunately, as is common with pipeline architecture, errors from OKB canonicalization would propagate to OKB linking in this case. Any noun phrase or relation phrase that was wrongly grouped via OKB canonicalization clearly cannot be linked correctly by the downstream OKB linking model. In fact, these two tasks are tightly coupled and one task can benefit significantly from the kind of information provided by the other. The idea of our joint OKB canonicalization and linking is based on two assumptions as follows:
	
	%\vspace{0.3mm}
	\textbf{\emph{Assumption 1:}} \textit{Two noun phrases (relation phrases) in OIE triples are more likely to be clustered to the same group if they are linked to the same entity (relation) in a CKB via OKB linking.}
	
	\textbf{\emph{Assumption 2:}} \textit{Two noun phrases (relation phrases) in OIE triples are more likely to be linked to the same entity (relation) in a CKB if they are clustered to the same group via OKB canonicalization.}
	%\vspace{0.3mm}

	%To promote knowledge transfer and fusion, researchers have put persistent effort into resolving the task of entity alignment (a.k.a. entity matching or entity resolution) by leveraging KG embedding techniques. The goal is to identify entities from different KBs that refer to the same real-world object while our task forcuses on the alignment between OKB and KB. Obviously, it can be seen that our task is relevant but different from this task and their models cannot be applied to address our task.
	
	%reduce space

	%It is observed that better OKB canonicalization result leads to better OKB linking result and vice versa. Meanwhile, errors made by one task may be corrected by another. Therefore, in this paper, we propose to jointly solve OKB canonicalization and OKB linking, which faces the following challenges.
	%\vspace{-1.3mm}
	%\begin{itemize}
	%	\item How to make these two tasks reinforce each other.
	%	\item How to make use of all useful signals from both tasks.
	%	\item How to make the framework flexible and able to extend to fit new signals of both tasks.
	%\end{itemize}

	It is observed that better OKB canonicalization result leads to better OKB linking result and vice versa. Meanwhile, errors made by one task may be corrected by the other. Therefore, in this paper, we propose to jointly solve OKB canonicalization and OKB linking, which faces the following challenges: (1) How to make these two tasks reinforce each other; (2) How to make use of all useful signals from both tasks; (3) How to make the framework flexible and able to extend to fit new signals of both tasks.
	%\vspace{-3mm}
	%\begin{itemize}[leftmargin=*]
	%\setlength{\itemsep}{0pt}
	%\setlength{\parsep}{0pt}
	%\setlength{\parskip}{0pt}
	%\setlength\parindent{0pt}
	%\item How to make these two tasks reinforce each other.
	%\item How to make use of all useful signals from both tasks.
	%\item How to make the framework flexible and able to extend to fit new signals of both tasks.
	%\vspace{-3mm}
	%\end{itemize}
	
	To address all the above issues, we propose a novel framework JOCL, which \underline{\textbf{J}}ointly solves \underline{\textbf{O}}KB \underline{\textbf{C}}anonicalization and \underline{\textbf{L}}inking based on factor graph model.
	Given a set of OIE triples in an OKB as well as a CKB, our framework firstly constructs a factor graph for each task respectively. With respect to the task of OKB canonicalization, JOCL generates a variable node for each pair of noun (relation) phrases to represent whether they have the same semantic meaning, and adds OKB canonicalization signals and transitive relation signals as factor nodes. With respect to the task of OKB linking, JOCL generates a variable node for each noun (relation) phrase to represent its corresponding entity (relation) in a CKB, and adds OKB linking signals and fact inclusion signals as factor nodes.
	Subsequently, in order to make these two tasks reinforce each other, consistency signals are added as factor nodes to perform the interaction between two tasks based on the two assumptions above.
	A reasonable working procedure (i.e., interact between two tasks after resolving them alone) is elaborately designed for the Loopy belief propagation (LBP) algorithm \cite{murphy1999loopy,kschischang2001factor,ran2018attention} to pass messages among different types of nodes on the factor graph and learn parameters of our framework. With the final messages, JOCL could compute the marginal probability for each node according to the learned weights, and infer the corresponding entity (relation) for each noun (relation) phrase and generate canonicalization groups of noun (relation) phrases jointly. It is noted that due to intrinsic characteristics of factor graph model, our framework JOCL is flexible to fit any new signals via adding suitable factor nodes.

	Our contributions can be summarized as follows:
	%and verifies the mutual reinforcement of resolving both tasks jointly.
	%\vspace{-1.4mm}
	%\begin{itemize}[leftmargin=*]
	%	\setlength{\itemsep}{0pt}
	%	\setlength{\parsep}{0pt}
	%	\setlength{\parskip}{0pt}
	%	\item jkljllllllllllllllllllllllllllllllllllllllllllkkkkkkkkkkkkkkkkkkkkkkkkkkkkkkkkkkkkkkkkkkkkkkkkkkkkkkkkkkk
	%	
	%\end{itemize}
	
	\setlist[itemize]{leftmargin=0mm,topsep=0mm,parsep=0mm,listparindent=0mm,itemindent=\parindent}
	
	\begin{itemize}[leftmargin=*]
		\setlength{\itemsep}{0pt}
		\setlength{\parsep}{0pt}
		\setlength{\parskip}{0pt}
		\setlength\parindent{0pt}
		\item We are the first to explore the task of joint OKB canonicalization and linking, a new and increasingly important problem due to its broad applications.
		
		\item We propose a novel framework JOCL to perform OKB canonicalization and linking jointly, and make them reinforce each other. JOCL is flexible enough to combine different signals from two tasks together, and able to extend to fit any new signals of both tasks based on factor graph model.
		
		\item A thorough experimental study over two real-world data sets shows that JOCL outperforms all the baseline methods for both tasks in terms of average F1 and accuracy.
		
	\end{itemize}
	%\vspace{-0.08mm}
	%The remainder of this paper is organized as follows. Section \ref{prelimi} introduces some preliminaries and notations. We describe the framework JOCL in Section \ref{JOCL}. Section \ref{experiment} presents the experimental results and Section \ref{related work} discusses the related work. Finally, we conclude this paper in Section \ref{conclustion}.
	
	\section{PRELIMINARIES AND NOTATIONS}
%	\textcolor{red}{In this section, we first briefly describe some notations and define our task. Then we give some basic concepts of the factor graph.}
	
	\label{prelimi}
	In a CKB, an entity is denoted by $e$, a relation is denote by $r$, the set of entities is denoted by $E$, and the set of relations is denoted by $R$. A fact in a CKB can be denoted by $<$$e_i$, $r_k$, $e_j$$>$, where $e_i$, $e_j$ $\in$ $E$ and $r_k$ $\in$ $R$.
	In an OKB, an OIE triple is denoted by $t_i$=$<$$s_i, p_i, o_i$$>$, where $s_i$ and $o_i$ are noun phrases (NPs) and $p_i$ is a relation phrase (RP). A set of OIE triples is denoted by $T=\{t_1,t_2,...\}$.
	
	%\begin{figure*}[!t]
	%	\centering
	%	\includegraphics[width=6.51in]{kuangjia}
	%	\vspace{-3mm}
	%	\caption{A summarization of the proposed framework JOCL.}
	%	\label{framework}
	%\end{figure*}

	\begin{definition}
		[\textbf{Joint OKB Canonicalization and Linking}] Given a set of OIE triples in an OKB and a CKB, the target of this joint task is to cluster NPs or RPs with the same semantic meaning into a group (i.e., OKB canonicalization) and meanwhile linking each group of NPs or RPs with their corresponding real world entity or relation in a CKB (i.e., OKB linking) jointly.\end{definition}

	For illustration, we show a running example of this task in Figure \ref{task}. The input is three OIE triples and a CKB. With respect to the canonicalization result (marked with blue ellipses), we should cluster NPs into four groups and cluster RPs into two groups. With respect to the linking result (marked with blue arrows), we should link each group of NPs or RPs with their corresponding entity or relation in the CKB.
	
	A factor graph consists of variable nodes, factor nodes, and edges. A variable node represents a random variable and a factor node represents a factor function among variable nodes. In our factor graph, we utilize exponential-linear functions to instantiate factor functions. A factor node and each of its related variable node are connected by an undirected edge. The two types of nodes in a factor graph form a bipartite and undirected graph. 

	\section{The framework: JOCL}
	\label{JOCL}
%	In this section, we introduce our framework JOCL in detail. 
The overall framework JOCL is shown in Figure \ref{framework}. We begin with the description of the factor graph for each task and then introduce the interaction between two tasks. Finally, we introduce the learning and inference algorithm of our framework.

	\subsection{OKB Canonicalization}
	OKB canonicalization consists of two subtasks, namely NP canonicalization and RP canonicalization.
	In this subsection we introduce the factor graph for the task of OKB canonicalization given a set of OIE triples in an OKB. We firstly introduce how to generate variable nodes and factor nodes for this task. Next, we introduce some useful signals of NP (RP) canonicalization and transitive relation, which are embedded in factor nodes.

	\subsubsection{Variable nodes}
	For any two OIE triples $t_i$ and $t_j$, three observed variable $s_{ij}$ ($p_{ij}$, $o_{ij}$) called subject (predicate, object) pair variables are used to represent the NP pair $(s_i, s_j)$, the RP pair $(p_i, p_j)$, and the NP pair $(o_i, o_j)$, respectively. Each observed variable has only one state as it is observed. We generate a variable node for each observed variable in the factor graph. Additionally, we define three canonicalization variables $x_{ij}$ ($y_{ij}$, $z_{ij}$) called subject (predicate, object) canonicalization variables, corresponding to the variables $s_{ij}$, $p_{ij}$, and $o_{ij}$ respectively. The canonicalization variable represents whether two NPs or RPs have the same semantic meaning. Therefore, each canonicalization variable has two states (i.e., 0 and 1). For example, $x_{ij}=1$ means NP $s_i$ and NP $s_j$ refer to the same entity. We generate a variable node for each canonicalization variable in the factor graph.

	\subsubsection{Factor nodes} We add six kinds of factor nodes: subject (predicate, object) canonicalization factor node $F_1$ ($F_2$, $F_3$), and subject (predicate, object) transitive relation factor node $U_1$ ($U_2$, $U_3$).
	
	We utilize some NP canonicalization signals to define $F_1$ ($F_3$) as a factor function over a subject (object) pair variable and its corresponding subject (object) canonicalization variable, and generate a subject (object) canonicalization factor node for this function in the factor graph. We utilize some RP canonicalization signals to define $F_2$ as a factor function over a predicate pair variable and its corresponding relation canonicalization variable, and generate a predicate canonicalization factor node for this function in the factor graph.
	We utilize transitive relation signals to define a factor function $U_1$ ($U_2$, $U_3$) over three subject (relation, object) canonicalization variables that satisfy transitive relations, and generate a transitive relation factor node for this function in the factor graph.

	\subsubsection{NP canonicalization signals}
	\label{npcans}
	
	\begin{itemize}[leftmargin=*]
		\setlength{\itemsep}{0pt}
		\setlength{\parsep}{0pt}
		\setlength{\parskip}{0pt}
		\item  IDF token overlap: Inverse document frequency (IDF) token overlap is based on the assumption that two NPs sharing infrequent words are more likely to refer to the same object in the world. 
	For example, it is likely that ``Warren Buffett'' and ``Buffett'' refer to the same entity which share an infrequent word ``Buffett''.
		In \cite{galarraga2014canonicalizing} this signal has been verified to be a very effective signal for canonicalization and we use it to calculate the similarity between two NPs denoted by $Sim_{idf}(s_i, s_j)$ as follows.
		\begin{equation*}
			\label{idf}
			Sim_{idf}(s_i,s_j)=\frac{\sum\nolimits_{x\in w(s_i)\cap w(s_j)}log(1+f(x))^{-1}}{\sum\nolimits_{x\in w(s_i)\cup w(s_j)}log(1+f(x))^{-1}}
		\end{equation*}
		where $w(\cdot)$ is the set of words of a string, and $f(x)$ is the frequency of the word $x$ in the collection
		of all words that appear in the NPs of
		the OIE triples. 
		%	For instance, the IDF token overlap similarity between ``" (i.e.,) and ``" (i.e.,) is.
		We define the feature function $f_{idf}$ based on $Sim_{idf}$ as follows.
		\begin{equation*}
			\label{IDF}
			f_{idf}(s_{ij},x_{ij})=\begin{cases}
				Sim_{idf}(s_i,s_j) \text{\ \ \  \ \ \ if $x_{ij}=1$}\\
				1-Sim_{idf}(s_i,s_j) \text{\ \ \ if $x_{ij}=0$}\\
			\end{cases}
		\end{equation*}
		%	For each word we obtain its pretrained word embedding \cite{joulin2017bag}.
		
		%We use the fastText model trained on Common Crawl and Wikipedia. we first map each word to a high-dimensional vector space
		%using fastText [28] pre-trained word embeddings.
		
		\item Word embedding: Word embeddings are the de-facto standard in language modeling and very popular in NLP. A word embedding maps words from a vocabulary to vectors of real numbers. Word embeddings are often learned from co-occurrences and neighborhoods of words in large corpora \cite{mikolov2013distributed, pennington2014glove}. The rationale is that the meaning of a word is captured by the contexts where it often appears, which is called ``distributional semantics''. For a NP which contains several words, we average the vectors of all the single words in the phrase as its embedding for simplicity. We use the cosine similarity to calculate the similarity between the embeddings of two NPs. The similarity between two NPs based on this signal can be denoted by $Sim_{emb}(s_i, s_j)$ and we define the feature function $f_{emb}$ based on it.
		\begin{equation*}
			\label{word embed}
			f_{emb}(s_{ij},x_{ij})=\begin{cases}
				Sim_{emb}(s_{i},s_{j}) \text{\ \ \  \ if $x_{ij}=1$}\\
				1-Sim_{emb}(s_{i},s_{j}) \text{\ \ \ if $x_{ij}=0$}\\
			\end{cases}
		\end{equation*}
	For instance, the score of $Sim_{emb}$(``Barack Obama'',  ``President Obama'') is $0.873$ using fastText \cite{grave2018learning} embeddings trained on Common Crawl via MindSpore Framework\footnote{https://www.mindspore.cn/en}, which indicates these two NPs are likely to refer to the same entity.
		
		\item PPDB: PPDB 2.0 \cite{pavlick2015ppdb} is a large collection of paraphrases in English. All the equivalent phrases are clustered into a group and each group is randomly assigned a representative. If two NPs have the same cluster representative according to the index, they are considered to be equivalent and we set value of similarity to $1$ otherwise $0$. The similarity between two NPs based on this signal can be denoted by $Sim_{PPDB}(s_i,s_j)$ and we define the feature function $f_{PPDB}$ based on it.
		\begin{equation*}
			\label{ppdb}
			f_{PPDB}(s_{ij},x_{ij})=\begin{cases}
				Sim_{PPDB}(s_i,s_j) \text{\ \ \  \ if $x_{ij}=1$}\\
				1-Sim_{PPDB}(s_i,s_j) \text{\ \ \ if $x_{ij}=0$}\\
			\end{cases}
		\end{equation*}
		%	Similarity, we can define as follows:
		%	\begin{equation}
		%	F_3(o_io_j, z_{ij})=\frac{1}{Z_3}exp\{\alpha^Tf1(o_io_j, z_{ij})\}	\end{equation}
		%	$Z_3=\sum\limits_{z_{ij}}exp\{\alpha^Tf1(o_io_j, z_{ij})\}$
%		For example, the score of $Sim_{PPDB}$(``management'',  ``administration'') is $1$, which indicates the semantic equivalence between these two NPs.
		
	\end{itemize}
	
	Lastly, we define $F_1$ based on all NP canonicalization signals.
	\begin{equation*}
		F_1(s_{ij}, x_{ij})=\frac{1}{Z_1}exp\{\boldsymbol\alpha_{1}^T\boldsymbol{f}_1(s_{ij}, x_{ij})\}, \
		Z_1=\sum\limits_{x_{ij}}exp\{\boldsymbol\alpha_{1}^T\boldsymbol{f}_1(s_{ij}, x_{ij})\}
	\end{equation*}
	where $\boldsymbol{f}_1$$=<$$f_{idf}, f_{emb}, f_{PPDB}$$>$ is a vector of feature functions;  $\boldsymbol\alpha_{1}$ denotes the corresponding weights of the feature functions. Similarly, we can define $F_3$ based on the NP canonicalization signals above as well.

	\subsubsection{RP canonicalization signals}
	\label{RPcanonicalizations}
	IDF token overlap, word embedding, and PPDB introduced above can be used as the RP canonicalization signals directly as well. Apart from these, inspired by \cite{vashishth2018cesi} we use the following two additional signals.
	
	\begin{itemize}[leftmargin=*]
		
		\item AMIE: AMIE algorithm \cite{galarraga2013amie} can judge whether two RPs represent the same semantic meaning by learning Horn rules. We take morphological normalized OIE triples as the input of AMIE, and the output of AMIE is a set of implication rules between two RPs $p_i$ and $p_j$ (e.g., $p_i \Rightarrow p_j$) based on statistical rule mining. If both $p_i \Rightarrow p_j$ and $p_j \Rightarrow p_i$ satisfy support and confidence thresholds, we consider two RPs (i.e., $p_i$ and $p_j$) have the same semantic meaning and set value of similarity to $1$ otherwise $0$. The similarity between two RPs using AMIE is denoted by $Sim_{AMIE}(p_i,p_j)$ and the feature function $f_{AMIE}$ can be defined based on it as follows.
		\begin{equation*}
			\label{ppdb}
			f_{AMIE}(p_{ij},y_{ij})=\begin{cases}
				Sim_{AMIE}(p_i,p_j) \text{\ \ \  \ if $y_{ij}=1$}\\
				1-Sim_{AMIE}(p_i,p_j) \text{\ \ \ if $y_{ij}=0$}\\
			\end{cases}
		\end{equation*}
		For instance, the score of $Sim_{AMIE}$(``is the capital of'', ``is the capital city of'') is $1$ on the OIE triple data set ReVerb45K in our experiments, which indicates these two RPs have the same semantic meaning.
		
		\item KBP: Stanford Knowledge
		Base Population (KBP) \cite{surdeanu2012multi} system can link a RP to a relation in a CKB. If the relations of two RPs fall in the same
			category, these two RPs are considered as equivalent and we set value of similarity to $1$ otherwise $0$. The similarity between two RPs using KBP can be denoted by $Sim_{KBP}(p_i,p_j)$ and the feature function $f_{KBP}$ can be defined based on it as follows.
		\begin{equation*}
			\label{KBP}
			f_{KBP}(p_{ij},y_{ij})=\begin{cases}
				Sim_{KBP}(p_i,p_j) \text{\ \ \  \ if $y_{ij}=1$}\\
				1-Sim_{KBP}(p_i,p_j) \text{\ \ \ if $y_{ij}=0$}\\
			\end{cases}
		\end{equation*}
		For example, the score of $Sim_{KBP}$(``was working at'', ``worked for'') is $1$, which indicates these two RPs have the same semantic meaning.
		%	\item WSD: WSD uses
		%	word-sense disambiguation \cite{banerjee2002adapted} with Wordnet \cite{miller1995wordnet} to identify possible synsets for a given NP. If two NPs
		%	share a common synset, then they are marked as similar. The similarity between two RPs using this signal can be denoted by $Sim_{WSD}(p_i,p_j)$ and the feature function $f_{WSD}$ can be denoted according it as follows.
		%	\begin{equation}
		%	\label{wordnet}
		%	f_{WSD}(p_{ij},y_{ij})=\begin{cases}
		%	Sim_{WSD}(p_i,p_j) \text{\ \ \  \ if $y_{ij}=1$}\\
		%	1-Sim_{WSD}(p_i,p_j) \text{\ \ \ if $y_{ij}=0$}\\
		%	\end{cases}
		%	\end{equation}
	\end{itemize}
	
	Lastly, we define $F_2$ based on all RP canonicalization signals.
	\begin{equation*}
		F_2(p_{ij},y_{ij})=\frac{1}{Z_2}exp\{\boldsymbol\alpha_2^T\boldsymbol{f}_2(p_{ij}, y_{ij})\}, \
		Z_2=\sum\limits_{y_{ij}}exp\{\boldsymbol\alpha_{2}^T\boldsymbol{f}_2(p_{ij}, y_{ij})\}
	\end{equation*}
	where $\boldsymbol{f}_2$$=$$<$$f_{idf}$, $f_{emb}$, $f_{PPDB}$, $f_{AMIE}$, $f_{KBP}$$>$ is a vector of feature functions; $\boldsymbol\alpha_2$ denotes the corresponding weights of the feature functions.

	\subsubsection{Transitive relation signals}
	We have the fact that the pairs satisfy transitive relations. As an example, if NP $s_1$ and NP $s_2$ have the same semantic meaning (i.e., canonicalization variable $x_{12}$$=$$1$), and $s_2$ and $s_3$ have the same semantic meaning (i.e., $x_{23}$$=$$1$), then we can deduce that $s_1$ and $s_3$ have the same semantic meaning (i.e., $x_{13}$$=$$1$). If values of this kind of three canonicalization variables satisfy transitive relations, they should be rewarded. If their values violate transitive relations, they should be penalized. Specifically, we define $u_1$ as a transitive relation feature function over three subject canonicalization variables $x_{ij}$, $x_{jk}$, and $x_{ik}$ that satisfy transitive relations. If values of all the three variables are 1 which case satisfies transitive relations, we will give a high score for $u_1$ heuristically. If only one of the three variables has a value of 0 and the other two are 1 which case violates transitive relations, we will give a low score heuristically. Otherwise, we will give a middle score. Scores range from $0$ to $1$. The high (middle, low) score is set to $0.9$ ($0.5$, $0.1$) in our experiments. We define the factor function $U_1$ as follows.
	\begin{equation*}
		U_1(x_{ij}, x_{jk}, x_{ik})=\frac{1}{N_{1}}exp\{\boldsymbol\beta_1 u_1(x_{ij}, x_{jk}, x_{ik})\}
	\end{equation*}
	where $N_{_1}$=$\sum\limits_{x_{ij}, x_{jk}, x_{ik}}exp\{\boldsymbol\beta_1 u_1(x_{ij}, x_{jk}, x_{ik})\}$; $\boldsymbol\beta_1$ denotes the corresponding weight.
	We can define $U_2$ ($U_3$) based on the predicate (object) canonicalization variables in a similar way.

	\subsection{OKB Linking}
	OKB linking consists of two subtasks, namely OKB entity linking and OKB relation linking. In this subsection we introduce the factor graph for the task of OKB linking given a set of OIE triples in an OKB and a CKB. We firstly introduce how to generate variable nodes and factor nodes for this task. Next, we introduce some useful signals of OKB entity (relation) linking and fact inclusion, which are embedded in factor nodes.

	%\subsubsection{Variable node}
	%Under the scenario of Joint Open Knowledge Base Canonicalization and Linking, all variables embeded in variable nodes in a factor graph is divided into two subsets, corresponding to the observed and hidden variables respectively.

	\subsubsection{Variable nodes}
	For an OIE triple $t_i$ in an OKB, we regard its NP $s_i$, RP $p_i$, and NP $o_i$ as three observed variables, namely subject variable, predicate variable, and object variable, respectively. Each observed variable only has one state as it is observed. We generate a variable node for each observed variable in the factor graph. Additionally, we define three linking variables $e_{s_i}$ ($r_{p_i}$, $e_{o_i}$) called subject (predicate, object) linking variables, corresponding to the observed variables $s_i$, $p_i$, and $o_i$, respectively. $e_{s_i}$ ($e_{o_i}$) represents the semantically corresponding entity existing in the CKB for NP $s_i$ ($o_i$) which has $|e_{s_i}|$ ($|e_{o_i}|$) possible states each of which is a candidate entity in the CKB that NP $s_i$ ($o_i$) may refer to. $r_{p_i}$ represents the semantically corresponding relation existing in the CKB for RP $p_i$. It has $|r_{p_i}|$ possible states each of which is a candidate relation in the CKB that $p_i$ may refer to. We generate a variable node for each linking variable in the factor graph.

	\subsubsection{Factor nodes} We add four kinds of factor nodes: subject (predicate, object) linking factor node $F_4$ ($F_5$, $F_6$), and fact inclusion factor node $U_4$.
	%\subsubsection{Factor nodes} In our factor graph, there are seven different types of factor nodes according the type of feature function: (1) Noun phrase Canonicalization factor function; (2) Relation phrase Canonicalization factor function; (3) OIE entity Linking factor function; (4) OIE relation linking factor function; (5) Transitivity factor function; (6) KB-based factor function; (7) Consistency factor function.
	
	We utilize some OKB entity linking signals to define $F_4$ ($F_6$) as a factor function over a subject (object) variable and its corresponding subject (object) linking variable, and generate a subject (object) linking factor node for this function in the factor graph. We utilize some OKB relation linking signals to define $F_5$ as a factor function over a predicate variable and its corresponding predicate linking variable, and generate a predicate linking factor node for this function in the factor graph. We utilize fact inclusion signals to define a factor function $U_4$ over a subject linking variable, a predicate linking variable, and an object linking variable that correspond to the same OIE triple, and generate a fact inclusion factor node for this function in the factor graph.
	%\subsection{Features}
	%
	%Inspired by some previous work about OIE cononicalization and OKB linking, we select some useful features and embeded them in the above seven factor nodes. We introduce them as follows:

	\subsubsection{OKB entity linking signals}
	
	\begin{itemize}[leftmargin=*]
		\setlength{\itemsep}{0pt}
		\setlength{\parsep}{0pt}
		\setlength{\parskip}{0pt}
		
		\item Entity popularity: The entity popularity is found to be very helpful in previous entity linking methods \cite{shen2014probabilistic,shen2018shine+, ran2018attention, guo2013link, hua2015microblog, hoffart2011robust}, which tells us the prior probability of the appearance of a candidate entity given an entity mention. Thus, we utilize entity popularity as a signal of OKB entity linking. We use anchor links in Wikipedia to calculate the popularity of a candidate entity given a NP and define the feature function $f_{pop}$ as follows.
	
		\begin{equation*}
			\label{pop}
			f_{pop}(s_i, e_{s_i})=\frac{count(s_i, e_{s_i})}{count(s_i)}
		\end{equation*}
		
		where count($s_i$) denotes the number of the subject $s_i$ occurring as the surface form of an anchor link in Wikipedia; count($s_i, e_{s_i}$) represents the number of anchor links with the surface form $s_i$ pointing to the candidate entity $e_{s_i}$. 
%		For instance, the score of $f_{pop}$(``obama'', ``barack  obama'') is $0.91$, which indicates the candidate entity ``barack obama'' is probably to be the mapping entity of NP ``obama'' without considering the context.
		
	\end{itemize}
	
	We also use word embedding and PPDB introduced in Section \ref{npcans} as other two OKB entity linking signals by computing  string similarity between surface forms of the subject $s_i$ and its candidate entity $e_{s_i}$. Specifically, their corresponding feature functions are defined as: $f'_{emb}(s_i, e_{s_i})=Sim_{emb}(s_i, e_{s_i})$ and $f'_{PPDB}(s_i, e_{s_i})=Sim_{PPDB}(s_i, e_{s_i})$. Lastly, we define $F_4$ based on all OKB entity linking signals.
	\begin{equation*}
		F_4(s_i, e_{s_i})=\frac{1}{Z_4}exp\{\boldsymbol\alpha_4^T\boldsymbol{f}_4(s_i,e_{s_i})\} ,\
		Z_4=\sum\limits_{e_{s_i}}exp\{\boldsymbol\alpha_{4}^T\boldsymbol{f}_4(s_{i}, e_{s_i})\}
	\end{equation*}
	where $\boldsymbol{f}_4=$$<$$f_{pop}$,$f'_{emb}$,$f'_{PPDB}$$>$ is a vector of feature functions;  $\boldsymbol\alpha_4$ denotes the corresponding weights of the feature functions. Similarly, we can define $F_6$ based on the OKB entity linking signals above as well.

	\subsubsection{OKB relation linking signals}
	Word embedding and PPDB introduce above can be used as the OKB relation linking signals directly as well. Besides these, we use the following two signals.
	
	\begin{itemize}[leftmargin=*]
		\setlength{\itemsep}{0pt}
		\setlength{\parsep}{0pt}
		\setlength{\parskip}{0pt}
		\item Ngram \cite{nakashole2013fine}: Ngram can convert a string into a set of ngrams (i.e., a sequence of \textit{n} characters). The similarity between strings based on ngram could be Jaccard similarity between their sets of ngrams. 
%	For example, the Ngram similarity between ``locate in'' (i.e., $p_1$ in Figure 1(a)) and ``location.contained by'' (i.e., $r_1$ in Figure 1(a)) is 0.7, which indicates the candidate relation $r_1$ may be the mapping relation of RP $p_1$.
		
		\item Levenshtein distance (LD): LD can calculate the number of deletions, insertions, or substitutions required to transform a string into another string, which could be regarded as the distance between strings. We normalize LD to a range from $0$ to $1$.
		
	\end{itemize}
	
	We define the feature functions $f_{ngram}(p_i, r_{p_i})$ and $f_{LD}(p_i, r_{p_i})$ by computing string similarity between surface forms of the predicate $p_i$ and its candidate relation $r_{p_i}$ via ngram and LD, respectively. We adopt a python library to compute those different string similarities in our experiments. Lastly, we define $F_5$ based on all OKB relation linking signals.

	\begin{equation*}
		F_5(p_i, r_{p_i})=\frac{1}{Z_5}exp\{\boldsymbol\alpha_5^T\boldsymbol{f}_5(p_i,r_{p_i})\}, \
		Z_5=\sum\limits_{r_{p_i}}exp\{\boldsymbol\alpha_{5}^T\boldsymbol{f}_5(p_{i}, r_{p_i})\}
	\end{equation*}
	%\vspace{-1mm}
	where $\boldsymbol{f}_5$=$<$$f_{ngram}$, $f_{LD}$, $f'_{emb}$,$f'_{PPDB}$$>$ is a vector of feature functions; $\boldsymbol\alpha_5$ defines the corresponding weights of the feature functions.

	\subsubsection{Fact inclusion signals}
	For an OIE triple, its corresponding entities and relation are likely to compose a triple already included in a CKB. Therefore, if values of linking variables with respect to an OIE triple $t_i$=$<$$s_i ,p_i , o_i$$>$ compose a triple already included in a CKB, they should be rewarded. Specifically, we define $u_{4}$ as a fact inclusion feature function over three linking variables $e_{s_i}$, $r_{p_i}$, and $e_{o_i}$ with respect to an OIE triple $t_i$=$<$$s_i ,p_i , o_i$$>$. If the triple $(e_{s_i}, r_{p_i}, e_{o_i})$ is a fact already included in a CKB, we will give a high score for $u_4$ heuristically otherwise a relatively low score. Scores range from $0$ to $1$. The high (low) score is set to $0.9$ ($0.1$) in our experiments. We define the factor function $U_4$ as follows.
	\begin{equation*}
		%\vspace{-2mm}
		U_4(e_{s_i}, r_{p_i}, e_{o_i})=\frac{1}{N_4}exp\{\boldsymbol\beta_4 u_4(e_{s_i}, r_{p_i}, e_{o_i})\}
	\end{equation*}
	where $N_4$=$\sum\limits_{e_{s_i}, r_{p_i}, e_{o_i}}$$exp\{\boldsymbol\beta_4 u_4(e_{s_i}, r_{p_i}, e_{o_i})\}$; $\boldsymbol\beta_4$ denotes the corresponding weight.
	\subsection{Interaction Between Two Tasks}
	
	To make two tasks
	%OKB canonicalization and OKB linking 
	reinforce each other, we utilize consistency signals to define a factor function $U_5$ ($U_6$, $U_7$) over a subject (predicate, object) canonicalization variable and its corresponding two subject (predicate, object) linking variables, and generate a consistency factor node for this function in the factor graph.
	
	Based on the two assumptions introduced in Section 1, we come to the conclusion that OKB canonicalization result and OKB linking result should be consistent. For example, if NP $s_1$ and NP $s_2$ are linked to the same entity in a CKB (i.e., $e_{s_1}$$=$$e_{s_2}$), they should have the same semantic meaning (i.e., $x_{12}$$=$$1$) and vice versa. If values of these corresponding variables satisfy consistency relations, they should be rewarded. Otherwise, they should be penalized. Specifically, we define $u_{5}$ as a consistency feature function over three variables $e_{s_i}$, $e_{s_j}$, and $x_{ij}$. If the value of $e_{s_i}$ equals the value of $e_{s_j}$ and the value of $x_{ij}$ is $1$, or the value of $e_{s_i}$ does not equal the value of $e_{s_j}$ and the value of $x_{ij}$ is $0$ which two cases satisfy consistency relations, we will give a high score for $u_5$ heuristically. Otherwise, we set a relatively low score heuristically. Scores range from $0$ to $1$. The high (low) score is set to $0.7$ ($0.3$) in our experiments. Then we define the factor function $U_5$ as follows.
	
	\begin{equation*}
		U_5(e_{s_i}, e_{s_j}, x_{ij})=\frac{1}{N_5}exp\{\boldsymbol\beta_5 u_5(e_{s_i}, e_{s_j}, x_{ij})\}
	\end{equation*}
	where $N_5$=$\sum\limits_{e_{s_i}, e_{s_j}, x_{ij}}$$exp\{\boldsymbol\beta_5 u_5(e_{s_i}, e_{s_j}, x_{ij})\}$; $\boldsymbol\beta_5$ denotes the corresponding weight.
	We can define $U_6$ ($U_7$) based on a predicate (object) canonicalization variable and its corresponding two predicate (object) linking variables in a similar way.

	\subsection{Learning}
	For our model, the factor function of any factor node can be represented in a unified form as $H_j$:

	\begin{equation}
	\label{hc}
	H_j(C_j)=\frac{1}{Z_j}exp\{ \boldsymbol{\omega}^T\boldsymbol{h}_j(C_j)
	\}
	\end{equation}
	where $j$ denotes a clique which is a fully connected subset of the variables in the graph, $C_j$ is the set of variable nodes in the clique $j$, $\boldsymbol{\omega} = (\omega_1, \omega_2, ...)$ is a weighting vector and $\boldsymbol{h}_j$ is a vector of feature functions. Learning a factor graph model is to estimate an optimum parameter configuration $\boldsymbol{\omega}^{*}$.
	
	According to the factorization principle
	in the factor graph \cite{kschischang2001factor}, we could use the product of these factor
	functions to represent the joint probability over variables as follows.
	\begin{equation}
	P(Y)=\frac{1}{Z}\prod_{j}H_j(C_j), \ Z=\sum\limits_{C_j}\prod_{j}H_j(C_j)
	\end{equation}
	
	where $Z$ is a normalization factor, which is the summation of all possible values for $C_j$. $Y$ is defined as a collection of variables in our framework as follows.
	\begin{equation}
	Y=\{e_{s_i},r_{p_i},e_{o_i},x_{ij},y_{ij},z_{ij},e_{s_j},r_{p_j},e_{o_j}\}
	\end{equation}
	where $t_i= < s_i ,p_i , o_i >$,  $t_j= < s_j ,p_j, o_j >$, and $t_i,t_j \in T$.
	We use $P(Y)$ as the object function of our task and rewrite this function according to Formula \ref{hc} as follows.
	\begin{equation}
	P(Y)=\frac{1}{Z'}exp\{\boldsymbol{\omega}^T\sum\limits_{j}h_j(C_j)\}=\frac{1}{Z'}exp\{\boldsymbol{\omega}^TQ\}
	\end{equation}
	where $Z'=\sum_{j}exp\{\boldsymbol{\omega}^TQ\}$ and $Q=\sum_{j}h_j(C_j)$. We can obtain the following log-likelihood objective function:
	\begin{equation}
	\begin{aligned}
	O(\boldsymbol\omega)&=logP(Y^L)=log\sum\limits_{Y|Y^L}\frac{1}{Z'}exp\{\boldsymbol{\omega}^TQ\}\\
	&=log\sum\limits_{Y|Y^L}exp\{\boldsymbol\omega^TQ\}-log\sum\limits_{Y}exp\{\boldsymbol\omega^TQ\}
	\end{aligned}
	\end{equation}
	where $Y^L$ denotes the known labels and $Y|Y^L$ is a labeling configuration of $Y$ inferred from $Y^L$. We use the gradient descent algorithm to maximize the objective function. The gradient for parameters $\boldsymbol\omega$ can be calculated as follows:

	\begin{equation}
	\begin{aligned}
	\frac{\partial{O(\boldsymbol\omega)}}{\partial\boldsymbol\omega}
	&=\frac{\partial{(log\sum_{Y|Y^L}exp\{\boldsymbol\omega^TQ\}-log\sum_{Y}exp\{\boldsymbol\omega^TQ\})}}{\partial\boldsymbol\omega}\\
	&=\frac{\sum_{Y|Y^L}exp\{\boldsymbol\omega^TQ\} \cdot{Q}}{\sum_{Y|Y^L}exp\{\boldsymbol\omega^TQ\}}-\frac{\sum_{Y}exp\{\boldsymbol\omega^TQ\}\cdot{Q}}{\sum_{Y}exp\{\boldsymbol\omega^TQ\}}\\
	&=\mathbb{E}_{p_{\boldsymbol\omega}(Y|Y^L)}Q-\mathbb{E}_{p_{\boldsymbol\omega}(Y)}Q
	\end{aligned}
	\end{equation}
	where $\mathbb{E}_{p_{\boldsymbol\omega}(Y|Y^L)}Q$ and $\mathbb{E}_{p_{\boldsymbol\omega}(Y)}Q$ are two expectations of $Q$ based on the probabilistic distribution $p_{\boldsymbol\omega}(Y|Y^L)$ and $p_{\boldsymbol\omega}(Y)$ respectively. To obtain the gradient, we need to calculate two marginal probabilities $p(y_i)$ and $p(y_i, y_j,y_k)$, where $y_i,y_j,y_k\in{Y}$. However, as the graph structure of the factor graph can be arbitrary and may contain cycles, we cannot calculate the exact marginal probabilities. We use a two-step LBP algorithm to approximate the marginal probabilities. Interested readers please refer to \cite{tang2011learning, tang2016transfer, ran2018attention} for details of the algorithm.
	
	Specifically, we design a reasonable working procedure for the LBP algorithm based on the structure characteristics of our factor graph. We conduct the process of the message passing from factor nodes to variable nodes as follows.
	
	\begin{itemize}[leftmargin=*]
		\setlength{\itemsep}{0pt}
		\setlength{\parsep}{0pt}
		\setlength{\parskip}{0pt}
		\item Update all the messages from canonicalization factor nodes to canonicalization variable nodes (i.e., $F_1$$\rightarrow$${x_{ij}}$, $F_2$$\rightarrow$${y_{ij}}$, $F_3$$\rightarrow$${z_{ij}}$).

		%	\begin{equation*}
		%		F_1\rightarrow{x_{ij}}\ \ F_2\rightarrow{y_{ij}}\ \ F_3\rightarrow{z_{ij}}
		%	\end{equation*}
		\item Update all the messages from transitive relation factor nodes to canonicalization variable nodes (i.e., $U_1$$\rightarrow$$x_{ij},x_{jk},x_{ik}\ \ $ $U_2$$\rightarrow y_{ij}$,$y_{jk}$,$y_{ik}\ \ $ $U_3$$\rightarrow$$z_{ij}$,$z_{jk}$,$z_{ik}$).
		%	\vspace{-1mm}
		%	\begin{equation*}
		%	U_1\rightarrow{x_{ij},x_{jk},x_{ik}}\ \ U_2\rightarrow{y_{ij},y_{jk},y_{ik}}\ \ U_3\rightarrow{z_{ij},z_{jk},z_{ik}}
		%	\end{equation*}
		%\vspace{-3mm}
		\item Update all the messages from linking factor nodes to linking variable nodes (i.e., $	F_4$$\rightarrow$$e_{s_i}$, $F_5$$\rightarrow$$r_{p_i}$, $F_6$$\rightarrow$$e_{o_i}$).
		%	\vspace{-1mm}
		%\begin{equation*}
		%	F_4\rightarrow{e_{s_i}}\ \ F_5\rightarrow{r_{p_i}}\ \  F_6\rightarrow{e_{o_i}}
		%\end{equation*}
		%\vspace{-4mm}
		\item Update all the messages from fact inclusion factor nodes to linking variable nodes (i.e., $U_4\rightarrow$$e_{s_i}, r_{p_i}, e_{o_i}$).
		%	\vspace{-1.5mm}
		%	\begin{equation*}
		%	U_4\rightarrow{e_{s_i}, r_{p_i}, e_{o_i}}
		%	\end{equation*}
		%\vspace{-5mm}
		\item Update all the messages from consistency factor nodes to canonicalization variable nodes and linking variable nodes (i.e., $U_5\rightarrow{e_{s_i},e_{s_j},x_{ij}}\ \ U_6\rightarrow{r_{p_i},r_{p_j},y_{ij}}\ \ U_7\rightarrow{e_{o_i},e_{o_j},z_{ij}}$).
		%	\begin{equation*}
		%	U_5\rightarrow{e_{s_i},e_{s_j},x_{ij}}\ \ U_6\rightarrow{r_{p_i},r_{p_j},y_{ij}}\ \ U_7\rightarrow{e_{o_i},e_{o_j},z_{ij}}
		%	\end{equation*}
	\end{itemize}
	%\vspace{-1.5mm}
	
	For the process of the message passing from variable nodes to factor nodes, we firstly update all the messages from  canonicalization variable nodes to each of their related factor nodes and then update all the messages from linking variable nodes to each of their related factor nodes.
	
	With the marginal probabilities,
	the gradient can be obtained by summing over all variables. In practice we found that convergence was achieved within twenty iterations. The learning algorithm also can be extended to a distributed learning version with a graph segmentation algorithm such as \cite{jo2018fast}.

	%\renewcommand{\algorithmicrequire}{\textbf{Input:}}
	%\renewcommand{\algorithmicensure}{\textbf{Output:}}
	%\begin{algorithm}[t]
	%	\caption{$The\ Learning\ Algorithm$}
	%	\label{a2}
	%	\begin{algorithmic}[1]
	%		\REQUIRE Factor graph, learning weight $\eta$.
	%		\ENSURE Feature weight vector $\boldsymbol\omega$.
	%		\medskip
	%		\STATE Initialize the feature weight vector $\boldsymbol\omega=\boldsymbol{0}$
	%		\REPEAT
	%		\STATE update all message $u$ according to the order ()
	%		
	%		\STATE update all message $v$ according to the order ($f_1-f_6$, $u_1-u_7$).
	%		\STATE $t=1$
	%		\STATE Initialize the feature weight vector $\overrightarrow{\omega}^{(t)}$ randomly
	%		\FOR{$n=1$ to $N$ }  %N
	%		\FOR{$i=1$ to $|E|$ }
	%		\STATE Compute $z^{e}$ for the \emph{i}-th entity $e$ in $E$ with current $\overrightarrow{\omega}^{(t)}$ via the linear function ranking model
	%		\STATE Update $\overrightarrow{\omega}^{(t)}$ by Formula \ref{deltaw}
	%		\STATE $t=t+1$
	%		\ENDFOR
	%		\ENDFOR
	%		\STATE $\overrightarrow{\omega}=\overrightarrow{\omega}^{(t-1)}$
	%		\UNTIL{the feature weight vector $\boldsymbol\omega$ stabilizes within some threshold}
	%	\end{algorithmic}
	%\end{algorithm}

	%We schedule these messages from entities to 3 to types and back rst. Next, we schedule messages from entities to 5 to
	%relations and back. Finally, from types to 4 to relations and
	%back. We repeat this schedule until message values converge
	%from one iteration to the next.
	
	\subsection{Inference}
%	\vspace{-1mm}
	After we learned the optimal parameters, we can infer the best
	label of each variable (i.e., the corresponding entity (relation) that each NP (RP) refers to, and whether two NPs or RPs represent the same semantic meaning) by computing the marginal probability for each node with the final messages. The best label could be the state with the highest marginal probability.
	
	Although the consistency signals tend to make the results of OKB canonicalization and OKB linking as consistent as possible, there are still some conflicts between them after the process of inference. To eliminate conflicts and generate the final result, we design an intuitive method. If a pair of NPs are located in two different groups according to the linking result and the corresponding canonicalization variable of this pair has a value of 1, we select the label of the larger group as the final label for both NPs.
	Finally, we will obtain canonicalization groups of NPs (RPs) and the corresponding entity (relation) in a CKB for each group of NPs (RPs).
	
	%\definecolor{Blue}{RGB}{65,105,225}
	
	%\definecolor{Blue}{RGB}{80,190,255}
	
	\definecolor{Blue}{RGB}{80,190,255}
	%\definecolor{Blue}{RGB}{100,200,255}
	%\definecolor{Blue}{RGB}{60,90,255}
	\begin{table*}[!t]
			\caption{Performance on the NP canonicalization task. All the results of the baselines are taken from SIST \cite{lin2019canonicalization}.}
			\label{npcan}
			\centering
			\resizebox{0.95\textwidth}{!}{
				\begin{tabular}{|c|ccc|c|ccc|c|}
					\hline
					\multirow{2}{*}{\textit{\textbf{Method}}} &
					\multicolumn{4}{c|}{\textit{\textbf{ReVerb45K}}} &
					\multicolumn{4}{c|}{\textit{\textbf{NYTimes2018}}}\\
					\cline{2-9}
					&\textit{\textbf{Macro F1}} &
					\textit{\textbf{Micro F1}}&
					\textit{\textbf{Pairwise F1}}&
					\textit{\textbf{Average F1}}
					&\textit{\textbf{Macro F1}} &
					\textit{\textbf{Micro F1}}&
					\textit{\textbf{Pairwise F1}}&
					\textit{\textbf{Average F1}}\\
					\hline
					\hline
					\makecell[c]{Morph Norm(2011) \cite{fader2011identifying}} & 0.281 & 0.699 & 0.653 & 0.544& 0.471& 0.658& 0.643& 0.591\\
					\hline
					\makecell[c]{Wikidata Integrator}  & 0.563 &  0.839  & 0.783  & 0.728& 0.476& 0.839& 0.783& 0.699\\
					\hline
					\makecell[c]{Text Similarity(2014) \cite{galarraga2014canonicalizing}}  & 0.543 &  0.821& 0.689 & 0.684& 0.581& 0.796& 0.658& 0.678\\
					\hline
					\makecell[c]{IDF Token Overlap(2014) \cite{galarraga2014canonicalizing}}  & 0.598  & 0.571 &  0.505 &  0.558 & 0.551& 0.612& 0.527& 0.563\\
					\hline
					\makecell[c]{Attriubte Overlap(2014) \cite{galarraga2014canonicalizing}}  & 0.598  & 0.599  & 0.587& 0.595& 0.551& 0.612& 0.527& 0.563\\
					\hline
					\makecell[c]{CESI(2018) \cite{vashishth2018cesi}}  & 0.618  & 0.845  & 0.819  & 0.761& 0.586& 0.842& 0.778& 0.735\\
					\hline
					\makecell[c]{SIST(2019) \cite{lin2019canonicalization}}  & \textbf{0.691} & 0.889 & 0.823 & 0.801& \textbf{0.675}& 0.816& 0.838& 0.776\\
					\hline
					\makecell[c]{JOCL}  & 0.684 & \textbf{0.892}  & \textbf{0.877}  & \textbf{0.818} & 0.561 & \textbf{0.921}& \textbf{0.934}& \textbf{0.805}\\
					\hline
				\end{tabular}
			}
	\end{table*}
	\begin{table}[!t]
		%\setlength{\abovecaptionskip}{5pt}
		%\setlength{\belowcaptionskip}{10pt}
		%	\begin{mdframed}[linecolor=blue,linewidth=0pt,innerrightmargin=23pt,innerbottommargin=0pt,innerleftmargin=23pt,innertopmargin=0pt]
		\caption{Performance on the RP canonicalization task. All the results of the baselines are taken from SIST \cite{lin2019canonicalization}.}
		\label{relationc}
		\centering
		\resizebox{0.47\textwidth}{!}
		{
			\begin{tabular}{|c|ccc|c|}
				\hline
				\multirow{1}{*}{\textit{\textbf{Method}}} & \textit{\textbf{Macro F1}} & \textit{\textbf{Micro F1}} & \textit{\textbf{Pairwise F1}} &\textit{\textbf{Average F1}}\\
				%			& \textit{\textbf{F1}} &\textit{\textbf{F1}} &\textit{\textbf{F1}}&\textit{\textbf{F1}}&\\
				\hline
				\hline
				
				AMIE(2013) \cite{galarraga2013amie} & 0.703 & 0.820 & 0.760&0.761\\
				\hline
				PATTY(2012) \cite{nakashole2012patty} & 0.782 & 0.872 & 0.802&0.819\\
				\hline
				SIST(2019) \cite{lin2019canonicalization}& \textbf{0.875} & 0.872 & 0.845&0.864 \\
				\hline
				JOCL&0.848&\textbf{0.923}&\textbf{0.851}&\textbf{0.874}\\
				\hline
			\end{tabular}
		}
		%\end{mdframed}
	\end{table}

	\section{EXPERIMENTAL STUDY}
%	\vspace{-1mm}
	\label{experiment}

	\subsection{Experiment Settings}
%	\vspace{-1mm}
	\label{setting}
	Two large scale publicly available OIE triple data sets are used in the experiments: ReVerb45K \cite{vashishth2018cesi} and NYTimes2018 \cite{lin2019canonicalization} which are common benchmark data sets for OKB canonicalization and OKB linking. The OIE triples of ReVerb45K are extracted by ReVerb \cite{fader2011identifying} from the source text in Clueweb09 and all NPs are annotated with their corresponding Freebase entities. ReVerb45K contains $45$K triples all associated with Freebase entities each of which has at least two aliases occurring as NP. The OIE triples of NYTimes2018 are extracted by Standford OIE Tool \cite{angeli2015leveraging} over $1500$ articles from nytimes.com in $2018$. NYTimes2018 contains $34$K triples which are not annotated with any CKB. In both data sets, no training set is given. We leverage the triples associated with $20\%$ selected Freebase entities of ReVerb45K as the validation set, and the rest triples of ReVerb45K and all the triples of NYTimes2018 as two test sets. In this experiment, we use the validation set to train the parameters of our framework, and the test set to evaluate the performance. 
	
	For the task of OKB canonicalization, we adopt the same evaluation measures (i.e., macro, micro, and pairwise metrics) as previous works \cite{galarraga2014canonicalizing,vashishth2018cesi,lin2019canonicalization}. Specifically, macro metric evaluates whether the NPs or RPs with the same semantic meaning have been clustered into a group, micro metric evaluates the purity of the resulting groups, and pairwise metric evaluates individual pairwise merging decisions. In each metric, F1 score is the harmonic mean of precision and recall. For the detailed computing methods of these metrics, we omit them due to limited space and you could refer to \cite{galarraga2014canonicalizing,vashishth2018cesi,lin2019canonicalization}. To give an overall evaluation of each method for OKB canonicalization, we calculate average F1 as the average of macro F1, micro F1, and pairwise F1, which is common practice in OKB canonicalization. For the evaluation measure of OKB linking, we adopt accuracy which is a common measure for entity linking systems \cite{shen2015entity} and calculated as the number of correctly linked NPs (RPs) divided by the total number of all NPs (RPs). As it is unnecessary and impractical to generate canonicalization variables for all pairs of NPs and RPs in the factor graph, we generate canonicalization variables only for NP (RP) pairs with a relatively high similarity based on IDF token overlap introduced in Section \ref{npcans}, whose threshold is set to $0.5$. The learning rate of the gradient descent algorithm is set to $0.05$ in all experiments. The source code and data sets used in this paper are publicly available\footnote{https://github.com/JOCL-repo/JOCL}.
	\label{effectiveness}
	%We evaluate the performance of JOCL on the NP canonicalization task, RL canonicalization task, OKB entity linking, and OKB relation linking against seven, two, four, and three state-of-the-art methods, respectively.
	%has been mentioned in Section \ref{npcans}
	\subsection{OKB Canonicalization Task}
	%In this subsection, we evaluate the performance of JOCL first on the NP canonicalization task against seven baselines on both data sets, and then on the RP canonicalization task against three baselines on Reverb45K.
	%\footnote{https://github.com/SuLab/WikidataIntegrator}
%	\vspace{-1mm}

	\subsubsection{NP canonicalization}
	\label{npcan4.2.1}
	
	The baselines are listed as follows.
	\begin{itemize}[leftmargin=*]
		\setlength{\itemsep}{0pt}
		\setlength{\parsep}{0pt}
		\setlength{\parskip}{0pt}
		\item Morph Norm \cite{fader2011identifying} uses some simple normalization operations (e.g., removing tenses, pluralization).
		\item Wikidata Integrator\footnote{https://github.com/SuLab/WikidataIntegrator} is an opensource entity linking tool. NPs linked to the same entity by it are grouped together.
		\item Text Similarity \cite{galarraga2014canonicalizing} calculates the similarity between two NPs by JaroWinkler \cite{winkler1999state} similarity and utilizes hierarchical agglomerative clustering (HAC) method.
		\item  IDF Token Overlap \cite{galarraga2014canonicalizing} calculates the similarity between two NPs based on IDF token overlap and utilizes HAC method for clustering.
		\item  Attribute Overlap \cite{galarraga2014canonicalizing} uses the Jaccard similarity of attributes between two NPs for canonicalization.
		\item CESI \cite{vashishth2018cesi} performs OKB canonicalization using learned embeddings and side information.
		\item  SIST \cite{lin2019canonicalization} is the state-of-the-art method for OKB canonicalization by leveraging side information from the original source text.
	\end{itemize}
	
	Specially, for NYTimes2018 data set which is not annotated with any CKB, we randomly sample 100 non-singleton NP groups and manually label them as the ground truth for NP canonicalization like SIST. From the results shown in Table \ref{npcan}, we can see that JOCL outperforms all the baselines in terms of average F1 on both data sets. CESI improves the quality of the NP canonicalization by using various side information (e.g., Wordnet, PPDB, and KBP). SIST outperforms CESI by leveraging more side information from the original source text, which puts JOCL at a disadvantage. However, in spite of this disadvantage, JOCL still promotes by about $1.7$ ($2.9$) percentages compared with
		SIST in terms of average F1 over ReVerb45K (NYTimes2018), which demonstrates the effectiveness of JOCL in NP canonicalization.
	%\subsubsection{Performance Analysis Of RP Canonicalization}

	\subsubsection{RP canonicalization}
	We utilize AMIE introduced in Section \ref{RPcanonicalizations}, PATTY \cite{nakashole2012patty}, and SIST as baselines of RP canonicalization over ReVerb45K. PATTY can put the triples with the same pairs of NPs, as well as RPs that belong to the same synset in PATTY in one group. We randomly sample 35 non-singleton RP groups and manually label them as the ground truth for RP canonicalization, which is the same as SIST. If two RPs have the same meaning after removing tense, pluralization, auxiliary verb, determiner, and modifier, they are considered to be the same. As shown in Table \ref{relationc}, compared with AMIE, our framework and other two baselines (i.e., PATTY and SIST) perform better, since the number of appearance for most RPs is less than the support threshold which leads AMIE only covers very few RPs. Compared with the state-of-the-art method SIST, JOCL promotes by about $1$ percentage in terms of average F1, which demonstrates the effectiveness of JOCL in RP canonicalization.
	
	%We utilize AMIE introduced in Section \ref{RPcanonicalizations}, PATTY \cite{nakashole2012patty}, and SIST as baselines of RP canonicalization over Reverb45K. PATTY can put the triples with the same pairs of NPs, as well as RPs that belong to the same synset in PATTY in one group. We randomly sample 35 non-singleton RP groups and manually label them as the ground truth for RP canonicalization like SIST. In addition to macro, micro, and pairwise F1, we show the RP groups generated by each method. As shown in Table \ref{relationc}, compared with AMIE, our framework and other two baselines (i.e., PATTY and SIST) perform better and induce a larger number of RP groups. Compared with the state-of-the-art method SIST, JOCL promotes by about $1$ percentage in terms of average F1-score, which demonstrates the effectiveness of JOCL in RP canonicalization.
		\begin{figure}[t] %需要subfigure宏包
		\centering
		\subfigcapskip=-8pt
		\begin{minipage}[t]{0.98\columnwidth}
			\raggedright  %居左
			\begin{minipage}[t]{.95\textwidth}
				\raggedright
				\begin{figure}[H]
					\centering
					\subfigure{   %修改图一小标题
						\begin{minipage}[h]{0.471\textwidth}
							%	\begin{mdframed}[linecolor=yellow,linewidth=3pt]
							\captionof{table}{Performance on OKB entity linking task.}
							\resizebox{\textwidth}{38pt}{
								\begin{tabular}{|c|c|c|c|c|c|c|c|}
									\hline
									\textit{\textbf{Method}} & $\boldsymbol{\mathit{ReVerb45K}}$ &
									$\boldsymbol{\mathit{NYTimes2018}}$
									\\
									\hline
									\hline
									$\rm Falcon$ & $0.541$ & 0.33\\
									\hline
									$\rm EARL$ & 0.473 & 0.25 \\
									\hline
									Spotlight & 0.716 & 0.26\\
									\hline
									$\rm Tagme$ & $0.316$ & 0.3 \\
									\hline
									$\rm KBPearl$ & $0.522$ & 0.46\\
									\hline
									$\rm JOCL$ & $\textbf{0.761}$ & \textbf{0.48} \\
									\hline
								\end{tabular}
							}
						\end{minipage}
						%\end{mdframed}
					}
					\subfigure
					{    %修改图二小标题
						\begin{minipage}[h]{.451\textwidth}
							\raggedleft
							\includegraphics[width=\textwidth]{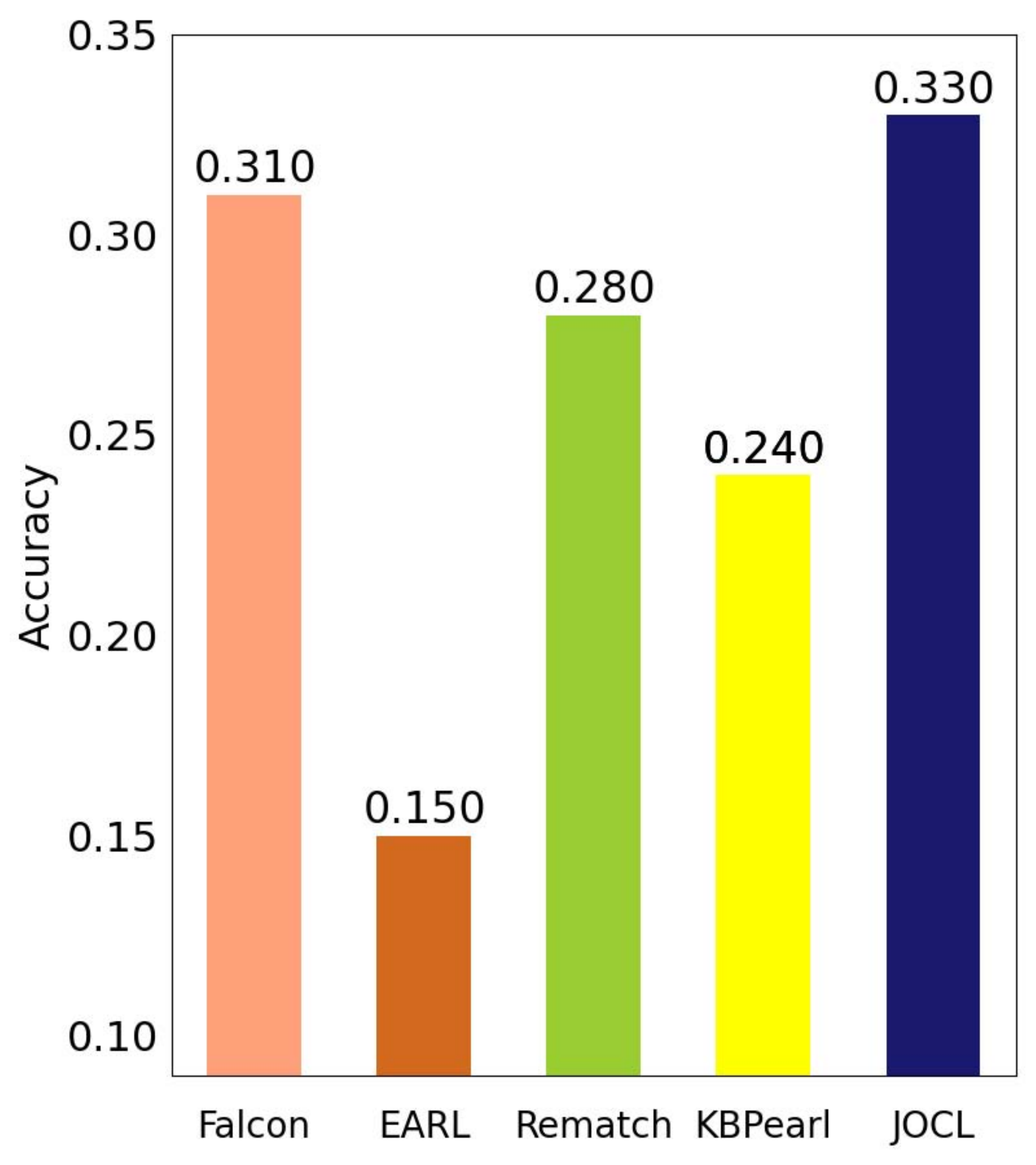}
							\caption*{Figure 3: Performance on OKB relation linking task.}
							%	\label{okblink2}
						\end{minipage}
					}
				\end{figure}
			\end{minipage}
		\end{minipage}
	\end{figure}
		\begin{table}
		\centering
		%		\begin{mdframed}[linecolor=blue,linewidth=0pt,innerrightmargin=0pt,innerbottommargin=0pt,innerleftmargin=0pt,innertopmargin=0pt]
		\caption{Performance of JOCL working separately for each task.}
		\label{onlytask}
		\centering
		\resizebox{0.47\textwidth}{!}{
			\begin{tabular}{|c|ccc|c|c|c|c|}
				\hline
				\textit{\textbf{Variant}} & $\boldsymbol{\mathit{Macro\ F1}}$ &
				$\boldsymbol{\mathit{Micro\ F1}}$ &
				$\boldsymbol{\mathit{Pairwise\ F1}}$ &
				$\boldsymbol{\mathit{Average\ F1}}$ & $\boldsymbol{\mathit{Accuracy}}$
				\\
				\hline
				\hline
				$\rm JOCL_{cano}$ & $0.571$ & $0.846$ &$0.787$ &$0.735$ &- \\
				\hline
				$\rm JOCL_{link}$ & - & -& -& -& $0.744$  \\
				\hline
				JOCL & \textbf{0.684} &\textbf{0.892} &\textbf{0.877} &\textbf{0.818} & \textbf{0.761}  \\
				\hline
			\end{tabular}
		}
		%		\end{mdframed}
		
	\end{table}

	\subsection{OKB Linking Task}
	\label{okblinkingtask}
	\subsubsection{OKB entity linking}
	The baselines are listed as follows.
	\begin{itemize}[leftmargin=*]
		\setlength{\itemsep}{0pt}
		\setlength{\parsep}{0pt}
		\setlength{\parskip}{0pt}
		\item Spotlight \cite{daiber2013improving, mendes2011dbpedia} is a popular baseline for entity linking based on DBpedia.
		\item TagMe \cite{ferragina2010tagme}
		is a popular baseline in the TAC KBP data sets for entity linking \cite{ji2014overview, raiman2018deeptype}.
		\item Falcon \cite{sakor2019old} performs joint entity and relation linking using some fundamental principles of English morphology.
		\item EARL \cite{dubey2018earl} performs joint entity and relation linking by solving a Generalized Traveling Salesman Problem (GTSP).
		\item KBPearl \cite{lin2020kbpearl} is a joint entity and relation linking system leveraging the context knowledge of the triples and side information inferred from the source text.
	\end{itemize}
	
	%	and uses a dictionary of entity surface
	%	forms extracted from Wikipedia to detect entity
	%	mentions in the parsed input text. These mentions
	%	passed through a voting scheme that computes the
	%	score for each mention-entity pair as the sum of
	%	votes given by candidate entities of all other mentions
	%	in the text (Ferragina and Scaiella, 2010),
	%	finally a pruning step filters out less relevant annotations.
	
	%	 and an extended CKB which merges entities and relations from DBpedia and Wikidata.

	%	Our approach models CKB relations
	%	with their underlying parts of speech, we then enhance this with
	%	extra aributes obtained from Wordnet and Dependency parsing
	%	characteristics. From a question, we model a similar representation
	%	of query relations. We then dene distance measurements between
	%	the query relation and the properties representations from the KG
	%	to identify which property is referred to by the relation within
	%	the query.
	%	
	%	is tool employs dependency
	%	parse characteristics with adjustment rules then carries out a match
	%	against KG properties enhanced with word lexicon Wordnet via a
	%	set of similarity measures.
	
	%	TagMe and Spotlight are the best two performing
	%	entity linking systems for question answering over DBpedia.
	Specially, for NYTimes2018 data set that is unlabeled, we randomly sample $100$ OIE triples and manually label each NP with its gold mapping entity as the ground truth for OKB entity linking task. For each baseline, we show its best performing result under its best parameter setting via running its different settings in Table 3. From the results on both data sets in Table 3, it can be seen that our framework JOCL outperforms all the five baselines on both data sets, which demonstrates the effectiveness of our framework for the task of OKB entity linking. 
	% Compared with the two best baselines Spotlight and Falcon, JOCL promotes by about $7.9$ percentages and $22.2$ percentages, respectively.
	
	%Given an Open IE KB, our rst goal is to canonicalize
	%its noun phrases. For simplicity, we concentrate here on
	%canonicalizing the subjects; the same approach can be used
	%to canonicalize the objects.
	
	%Specifically, some of the tools (DBpedia Spotlight, TagMe,
	%Falcon, and EARL) are designed to process short text. To
	%make them feasible on long-text documents, we follow the
	%setting in TagMe [20] to conduct sentence tokenization on
	%the documents with more than 30 words. One document
	%is divided into a list of sentences to pass to these tools for
	%further processing.

	\subsubsection{OKB relation linking}
	Falcon, EARL, and KBPearl can be used as the baselines of OKB relation linking as well. Apart from these, we use Rematch \cite{mulang2017matching} which is one of the top-performing tools for the relation linking task as a baseline. We randomly sample $100$ OIE triples of ReVerb45K and manually label each RP as the ground truth for OKB relation linking task. For each baseline, we show its best performing result under its best parameter setting in Figure 3. It can be seen that JOCL outperforms all the four baselines. The performance of
	all the methods on this task is not well compared with the OKB entity linking task, since the relations expressed in the OIE triples have much more representations than entities which makes this task very challenging.
	% which demonstrates the effectiveness of JOCL on OKB relation linking.
	%\begin{itemize}
	%	\item ReMatch
	%\end{itemize}
	
	%\end{mdframed}

		\subsection{Effect Analysis of Interaction Between Two Tasks}
		\label{interaction effect}
		To verify the effectiveness of interaction between two tasks, we remove consistency factor nodes introduced in Section 3.3 from JOCL and make these two tasks unable to interact with each other. We present the performance of two variants, namely, $\rm JOCL_{cano}$ (i.e., JOCL working on OKB canonicalization task alone) and $\rm JOCL_{link}$ (i.e., JOCL working on OKB linking task alone), as well as the whole framework JOCL on ReVerb45K in Table \ref{onlytask}. From the experimental results, we can see that JOCL outperforms $\rm JOCL_{cano}$ ($\rm JOCL_{link}$) in terms of average F1 (accuracy) over OKB canonicalization (linking) task, which demonstrates that the whole framework JOCL could indeed perform the interaction between two tasks effectively and make them reinforce each other obviously. For the canonicalization task, $\rm JOCL_{cano}$ outperforms most baselines except CESI and SIST in terms of average F1 shown in Table 1, since these two outperforming baselines use extra context knowledge inferred from the source text. For the linking task, $\rm JOCL_{link}$ outperforms all the baselines in terms of accuracy shown in Table 3.

	\begin{figure} %需要subfigure宏包
		\centering
		%		\subfigcapskip=-1pt
		\begin{minipage}[t]{1.02\columnwidth}
			%\raggedleft   %居右
			\begin{minipage}[t]{.95\textwidth}
				%\raggedleft
				\begin{figure}[H]\setcounter{subfigure}{0}
					\centering
					\subfigure[NP canonicalization.]{
						\begin{minipage}[h]{.451\textwidth}
							\raggedright
							\includegraphics[width=\textwidth]{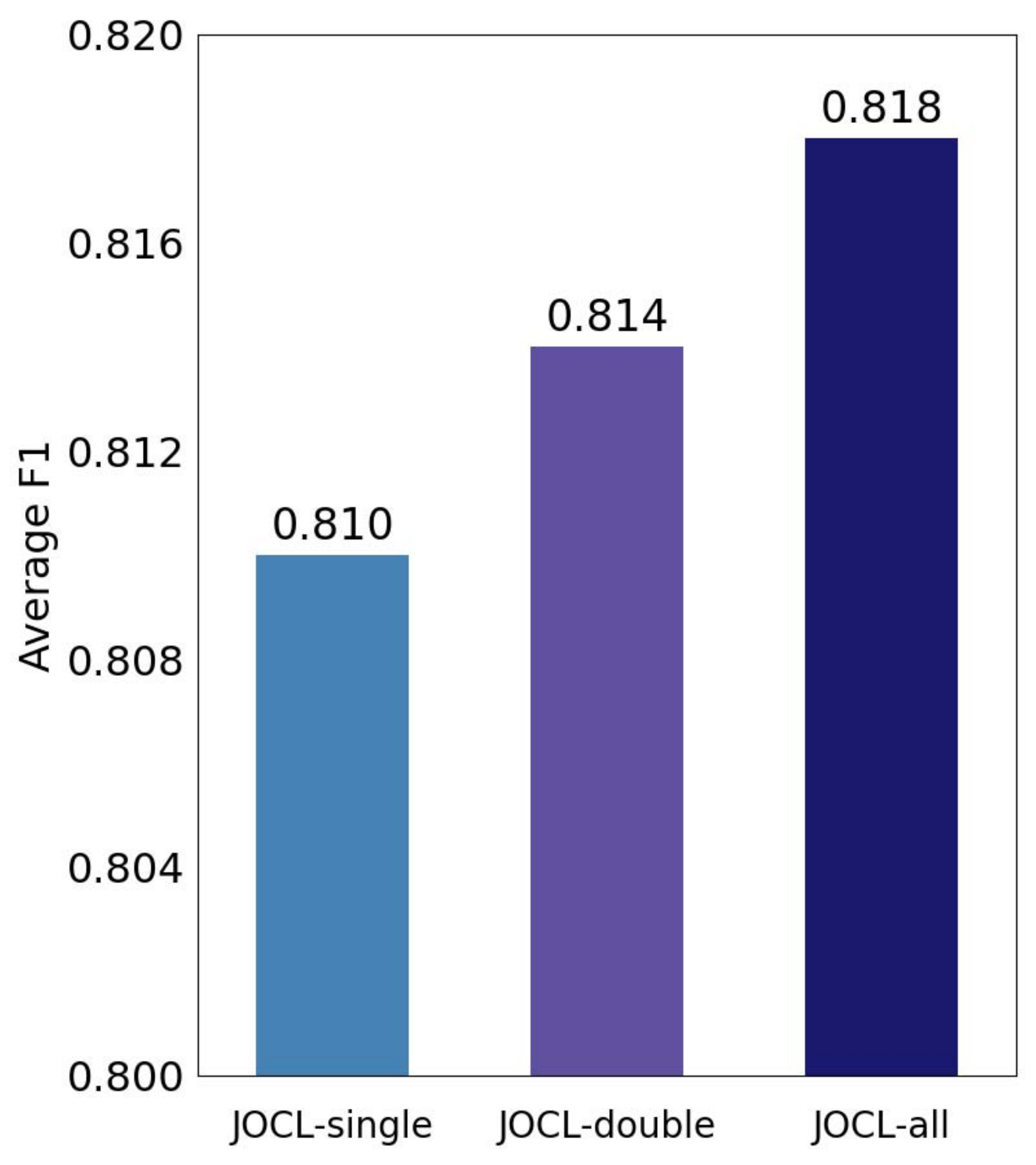}
						\end{minipage}
					}
					\subfigure[OKB entity linking.]{
						\begin{minipage}[h]{.451\textwidth}
							\raggedleft
							\includegraphics[width=\textwidth]{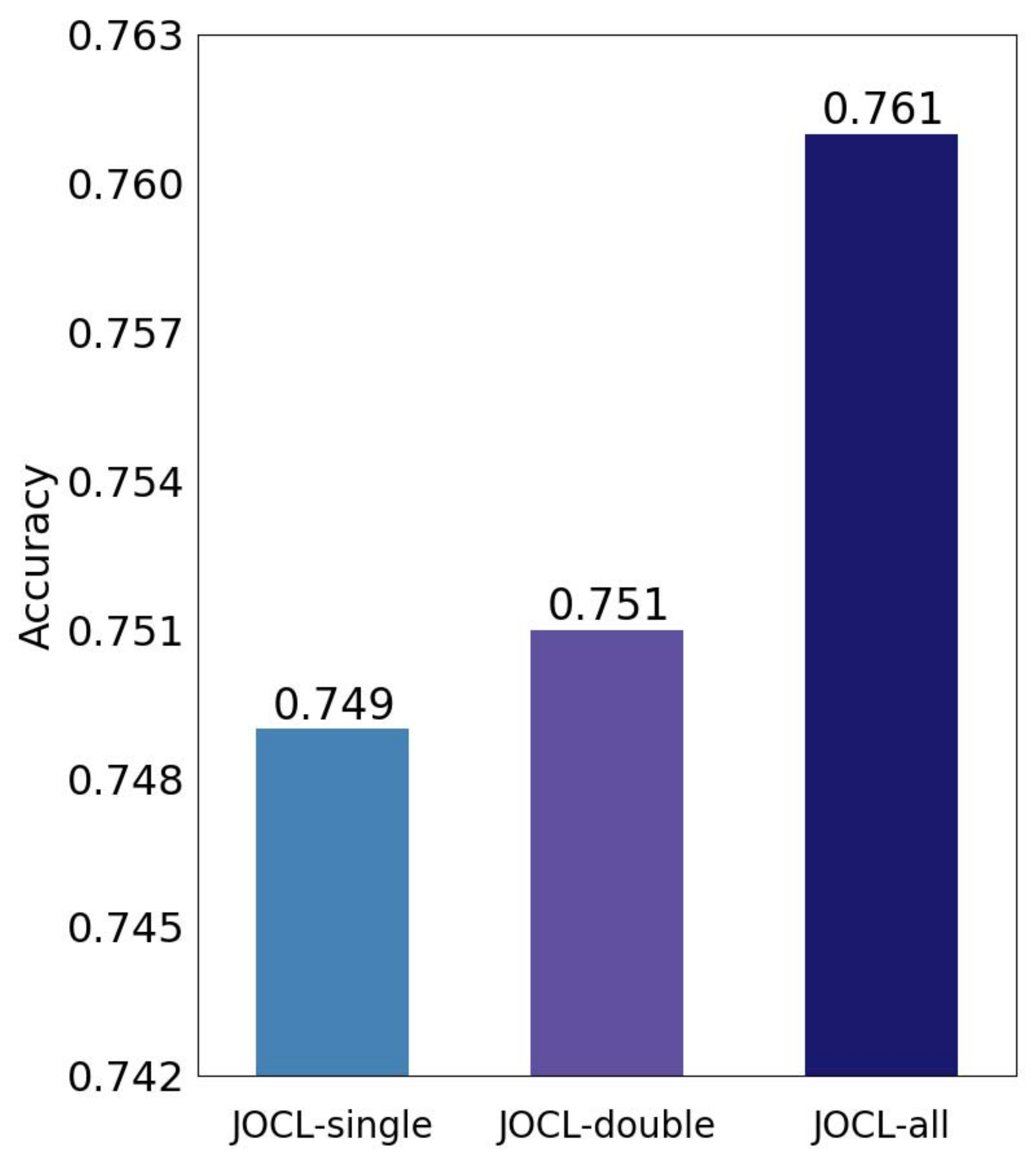}
						\end{minipage}
					}
					\caption*{Figure 4: Performance of different variants of JOCL for feature effect analysis.}
%					\vspace{-11mm}
				\end{figure}
			\end{minipage}
		\end{minipage}
	\end{figure}

	\subsection{Effect Analysis of Different Combinations of Feature Functions}
	\label{featuregroup}
	We define three different variants of our framework JOCL (i.e., JOCL-single, JOCL-double, JOCL-all) shown in Table \ref{configurations} by leveraging different combinations of the feature functions for each factor function in the factor graph, and present their performance on the NP canonicalization task (i.e., Figure 4(a)) and on the OKB
	entity linking task (i.e., Figure 4(b)) over ReVerb45K, respectively. From the experimental results, we can see that JOCL-all leveraging all the feature functions of each factor function achieves the best performance for both tasks. The more useful signals, the better the performance, which is consistent with our intuition. Moreover, we can see that JOCL is flexible enough to combine different signals from two tasks and able to extend to fit any new signals.
%	 \textcolor{blue}{Specially, it can be seen that IDF token overlap, word embedding, entity popularity, and Ngram are core signals  in comparison with other signals for our task.}
	
	\begin{table}
	\centering
	%		\vspace{-2mm}
	%	\caption{Variants of JOCL for feature effect analysis.}
	\caption{Different variants of JOCL for feature effect analysis.}
	\label{configurations}
	\resizebox{0.47\textwidth}{!}{
		\begin{tabular}{|c|c|c|c|c|}
			\hline
			\textit{\textbf{Variant}} & $\boldsymbol{\mathit{F_1}}$, $\boldsymbol{\mathit{F_3}}$ & $\boldsymbol{\mathit{F_2}}$ & $\boldsymbol{\mathit{F_4}}$, $\boldsymbol{\mathit{F_6}}$ & $\boldsymbol{\mathit{F_5}}$ \\
			\hline
			\hline
			JOCL-single & $f_{idf}$ & $f_{idf}$  & $f_{pop}$ & $f_{ngram}$ \\
			\hline
			JOCL-double & $f_{idf}, f_{emb}$ & $f_{idf}, f_{emb}$  & $f_{pop}, f'_{emb}$ & $f_{ngram},f'_{emb}$\\
			\hline
			JOCL-all & $\boldsymbol{f}_1$ & $\boldsymbol{f}_2$ & $\boldsymbol{f}_4$ & $\boldsymbol{f}_5$ \\
			\hline
		\end{tabular}
	}
\end{table}

	%\begin{table*}
	%  \caption{Some Typical Commands}
	%  \label{tab:commands}
	%  \begin{tabular}{ccl}
	%    \toprule
	%    Command &A Number & Comments\\
	%    \midrule
	%    \texttt{{\char'134}author} & 100& Author \\
	%    \texttt{{\char'134}table}& 300 & For tables\\
	%    \texttt{{\char'134}table*}& 400& For wider tables\\
	%    \bottomrule
	%  \end{tabular}
	%\end{table*}
	\section{related work}
	\label{related work}
	Four aspects of research are related to our work: OKB canonicalization, OKB linking, factor graph model, and entity resolution, which are introduced in detail as follows.

	The first work for OKB canonicalization \cite{galarraga2014canonicalizing} clusters NPs using some manually-defined signals to obtain equivalent NPs, and clusters RPs based on rules discovered by AMIE \cite{galarraga2013amie}. FAC \cite{wu2018towards} proposes a more efficient graph-based clustering method by pruning and bounding techniques.
	CESI \cite{vashishth2018cesi} learns the embeddings of NPs and RPs leveraging side information in a principled manner and clusters the learned embeddings together to obtain canonicalized NP (RP) groups. SIST \cite{lin2019canonicalization} proposes to leverage side information from the original source text (i.e., candidate entities of NPs, types of candidate entities, and the domain knowledge of the source text) to further improve the OKB canonicalization result.

	%Tang et al. [29] studied the social relationship classification problem on a publication dataset.Wang et al. [30] studied the cross-lingual knowledge linking problem on a dataset constructed from Wikipedia. The datasets they used are both large-scale.
	
	The most related work to OKB linking is the task of joint entity and relation linking. EARL \cite{dubey2018earl} designs two solutions: (1) formalizes the joint entity and relation linking tasks as an instance of the GTSP; (2) exploits the connection density between nodes in the graph. Falcon \cite{sakor2019old} utilizes the fundamental principles of English morphology (e.g., compounding and headword identification) and an extended knowledge graph created by merging entities and relations from various knowledge sources to capture semantics underlying the text.
	KBPearl \cite{lin2020kbpearl} performs joint entity and relation linking utilizing the context knowledge of the triples and side information extracted from the source documents. KBPearl relies on the source document, while our JOCL only focuses on the OIE triples without requiring the original source text.
	
	%However, these linking methods are not applicable to OIE triples which may contain new entities which are not exist in an CKB.
	%
	%Due to the ambiguity in OIE triples data set, models above performs poorly for the task of OKB linking.

	%they firstly canonicalize OIE triples. And they use the groups that were generated by OKB canonicalization as the input for the OKB linking. Unfortunately, as is common with pipeline architectures, errors from the OKB canonicalization propagate to the OKB linking. Any noun phrase or relation phrase that is wrongly grouped via OKB canonicalization clearly cannot be linked correctly by the downstream OKB linking model.

	The factor graph model has been successfully applied in many applications, such as knowledge base alignment \cite{wang2012cross}, social relationship mining \cite{tang2011learning, tang2016transfer}, social influence analysis \cite{tang2009social}, Web table annotation\cite{limaye2010annotating, zhang2013infogather+}, co-investment of venture capital \cite{wang2015prediction}, relationship prediction in E-commerce platform \cite{cen2019trust}, and tweet entity linking \cite{ran2018attention}. In this paper, we apply the factor graph model to jointly solve OKB canonicalization and OKB linking successfully.
	
	The task of entity resolution \cite{elmagarmid2006duplicate,getoor2012entity, mudgal2018deep} is related but different from our task. In entity resolution, each record describing an entity that needs to be matched contains a set of attribute values of this entity, while in our task the NPs and RPs that need to be clustered and linked reside in OIE triples and do not have attribute values with them.

	\section{conclusion}
	\label{conclustion}
	%OKB canonicalization and OKB linking are tightly coupled tasks, however, they have been studied in isolation so far. This paper is the first work to solve both tasks jointly. A whole framework JOCL based on factor graph proposed by us can make these two tasks reinforce each other. JOCL can combine signals from two tasks and extend to fit new signals.
	%To demonstrate the effectiveness of JOCL, we conduct experiments over a popular large scale OIE triples data set. Experimental results show that our framework outperforms all the baselines for both tasks. We are planning to release the source code upon the paper published.
	OKB canonicalization and OKB linking are tightly coupled tasks, and one task can benefit significantly from the other. However, previous studies only focus on one of them and cannot solve them jointly. To achieve this goal, we propose a novel framework JOCL based on factor graph model to perform interaction between two tasks and make them reinforce each other. JOCL can combine signals from both tasks and extend to fit new signals. To demonstrate the effectiveness of JOCL, we conduct experiments over two large scale OIE triple data sets and the experimental results show that our framework outperforms all the baselines for both tasks in terms of average F1 and accuracy. 
	
%	\textcolor{blue}{In general, however, the grouping and linking will be partial, meaning that some coreferences may be missed and some of the coreference groups may still be unlinkable – either because of remaining uncertainty or because the proper entity does not exist in the KB.}
%	
	%\newpage

	\section{Acknowledgments}
%	This work was supported in part by National Natural Science Foundation of China under Grant No. U1936206 and 61772289, Natural Science Foundation of Tianjin under Grant No. 19JCQNJC00100, YESS by CAST under Grant No. 2019QNRC001, and CAAI-Huawei MindSpore Open Fund. Jianyong Wang was supported in part by National Key Research and Development Program of China under Grant No. 2020YFA0804503, National Natural Science Foundation of China under Grant No. 61532010 and 61521002, and Beijing Academy of Artificial Intelligence (BAAI).
	This work was supported in part by National Natural Science Foundation of China (No. U1936206, 61772289), Natural Science Foundation of Tianjin (No. 19JCQNJC00100), YESS by CAST (No. 2019QNRC001), and CAAI-Huawei MindSpore Open Fund. Jianyong Wang was supported in part by National Key Research and Development Program of China (No. 2020YFA0804503), National Natural Science Foundation of China (No. 61532010, 61521002), and Beijing Academy of Artificial Intelligence (BAAI).

	\bibliographystyle{ACM-Reference-Format}
	 
	\bibliography{sample-base}

\end{document}